\pdfoutput=1

\documentclass[11pt]{article}

\usepackage[final]{acl}

\usepackage{times}
\usepackage{latexsym}

\usepackage[T1]{fontenc}

\usepackage[utf8]{inputenc}

\usepackage{microtype}

\usepackage{inconsolata}

\usepackage{graphicx}

\usepackage{xcolor}
\usepackage{listings}
\lstdefinelanguage{json}{
    basicstyle=\ttfamily\small,
    numbers=left,
    numberstyle=\scriptsize,
    stepnumber=1,
    numbersep=8pt,
    showstringspaces=false,
    breaklines=true,
    frame=single,
    backgroundcolor=\color{gray!10},
    literate=
     *{0}{{{\color{blue}0}}}{1}
      {1}{{{\color{blue}1}}}{1}
      {2}{{{\color{blue}2}}}{1}
      {3}{{{\color{blue}3}}}{1}
      {4}{{{\color{blue}4}}}{1}
      {5}{{{\color{blue}5}}}{1}
      {6}{{{\color{blue}6}}}{1}
      {7}{{{\color{blue}7}}}{1}
      {8}{{{\color{blue}8}}}{1}
      {9}{{{\color{blue}9}}}{1}
      {:}{{{\color{red}:}}}{1}
      {,}{{{\color{red},}}}{1}
      {\{}{{{\color{black}{\{}}}}{1}
      {\}}{{{\color{black}{\}}}}}{1}
      [{{{\color{black}{[}}}}{1}
      ]{{{\color{black}{]}}}}{1}
}
\usepackage{afterpage}

\usepackage{amsmath}
\usepackage{booktabs} 
\usepackage{hyperref}
\usepackage{framed}
\usepackage[utf8]{inputenc}
\usepackage{fancyhdr}
\newcommand{\ours}{\textsc{DIAMOND}}

\title{\ours: An LLM-Driven Agent for \\ Context-Aware Baseball Highlight Summarization}

\author{
  Jeonghun Kang \\
  TVING \\
  \And
  Soonmok Kwon \\
  TVING \\
  \And
  Joonseok Lee \\
  Seoul National University \\
  \And
  Byung-Hak Kim\thanks{Corresponding author: \texttt{bhak.kim@cj.net}} \\
  CJ Corporation \\
}

\begin{document}
\maketitle

\maketitle

\begin{abstract}
Highlight summarization in baseball requires balancing statistical analysis with narrative coherence. Traditional approaches—such as Win Probability Added (WPA)-based ranking or computer vision-driven event detection—can identify scoring plays but often miss strategic depth, momentum shifts, and storyline progression. Manual curation remains the gold standard but is resource-intensive and not scalable.
We introduce \textbf{\ours}, an \textbf{LLM-driven agent for context-aware baseball highlight summarization} that integrates \textbf{structured sports analytics with natural language reasoning}. \ours{} leverages sabermetric features—Win Expectancy, WPA, and Leverage Index—to quantify play importance, while an LLM module enhances selection based on contextual narrative value. This hybrid approach ensures both \textbf{quantitative rigor and qualitative richness}, surpassing the limitations of purely statistical or vision-based systems.
Evaluated on five diverse Korean Baseball Organization League games, \ours{} improves F1-score from 42.9\% (WPA-only) to 84.8\%, outperforming both commercial and statistical baselines. Though limited in scale, our results highlight the potential of modular, interpretable agent-based frameworks for event-level summarization in sports and beyond.
\end{abstract}

\section{Introduction}

Automating sports highlight generation presents a unique AI challenge: identifying impactful moments while preserving narrative coherence. In fast-paced sports like soccer or basketball, where crowd reactions and visual intensity often align with highlight-worthy moments, traditional AI methods—based on visual/audio cues—have been successful. In contrast, baseball presents a more episodic structure, where pivotal plays may be subtle and strategic rather than visually dramatic. 

Traditional highlight generation methods rely on visual, audio, or social signals. Visual-based models capture events like home runs or diving catches~\cite{joshi2017ibm, shih2017survey, merler2018automatic}, but miss strategic plays such as defensive shifts or base-running decisions. Audio-based methods use crowd noise or commentator emphasis~\cite{fu2017video, jiang2020towards}, which may reflect excitement but not analytical significance. Social engagement spikes~\cite{bettadapura2016leveraging} often correlate with popular moments but not with game-critical context. These methods overlook the nuance of narrative flow and strategic buildup, particularly in sports like baseball.

\begin{figure*}[t]
    \centering
    \includegraphics[width=0.8\textwidth]{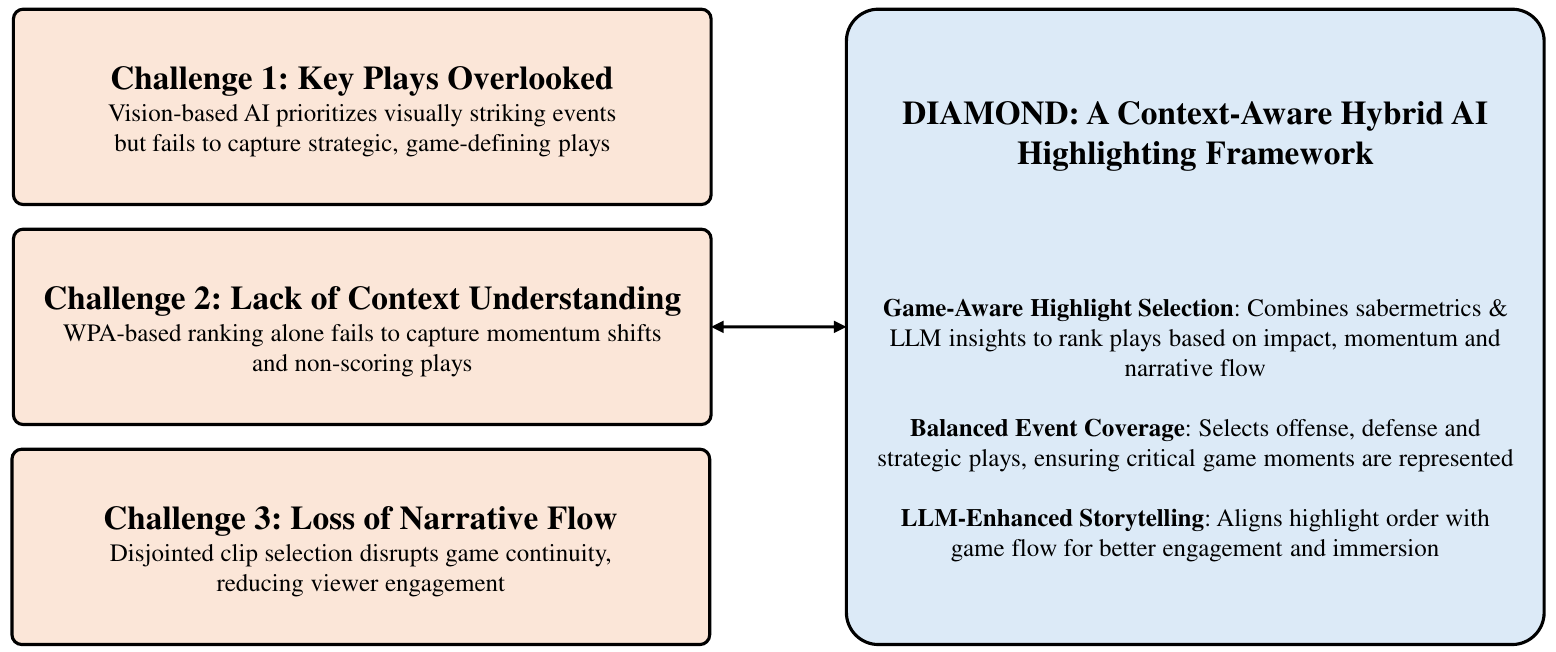}
    \caption{\textbf{Challenges in baseball highlight generation} include missed strategic plays, reliance on surface cues, and narrative loss. \ours{} addresses these by combining sabermetrics with LLM-driven contextual analysis for more engaging and data-driven highlight selection.}
    \label{fig:diamond_challenges}
\end{figure*}

Sabermetrics offers a structured, domain-specific lens for evaluating play impact through statistics like \textbf{Win Expectancy (WE)}, \textbf{Win Probability Added (WPA)}, and \textbf{Leverage Index (LI)}~\cite{tango2007book, fangraphs_we, retrosheet}. While useful for ranking events by outcome influence, purely statistical metrics fail to capture storytelling elements like momentum, context, or emotional resonance. This motivates approaches that go beyond quantification to capture qualitative game dynamics.

We introduce \textbf{\ours}, an \textbf{LLM-driven agent for context-aware baseball highlight summarization}. The name \ours{} evokes both the baseball field ("diamond") and the framework’s goal of producing high-fidelity, interpretable highlights. Figure~\ref{fig:diamond_challenges} summarizes the challenges DIAMOND addresses through a structured three-stage pipeline: Preparation, Decision, and Reflection. In the \textbf{Preparation Stage}, we process structured play-by-play logs and compute sabermetric scores. In the \textbf{Decision Stage}, we incorporate LLM-based contextual scoring to assess narrative relevance. Finally, in the \textbf{Reflection Stage}, we apply user-defined preferences and adjust rankings for narrative coherence.

\ours{} is designed to be \textbf{modular and domain-agnostic}. While our experiments are scoped to baseball, the framework generalizes to other sports or sequential domains (e.g., soccer, esports, financial news) by substituting appropriate domain-specific metrics. Its language-based approach avoids reliance on visual/audio data, offering scalability, interpretability, and ease of integration into real-world workflows.

Our key contributions are as follows:
\begin{itemize}
    \item We propose \ours{}, a modular agent framework that combines structured sabermetric scoring with LLM-driven contextual reasoning to generate baseball highlights that are both statistically grounded and narratively coherent.
    \item We introduce a three-stage agent pipeline—Preparation, Decision, and Reflection—that integrates quantitative and qualitative analysis while supporting user customization.
    \item We validate \ours{} through a mixed-method evaluation on five Korean Baseball Organization (KBO) League games, demonstrating substantial F1-score improvements over both commercial and statistical baselines.
\end{itemize}

By bridging structured sports analytics with LLM-based contextual reasoning, \ours{} offers an interpretable, agent-like framework for highlight generation—connecting event-level summarization with broader goals in NLP, AI-assisted media, and domain-aware agent design.

\begin{figure*}
    \centering
    \includegraphics[width=1.0\textwidth]{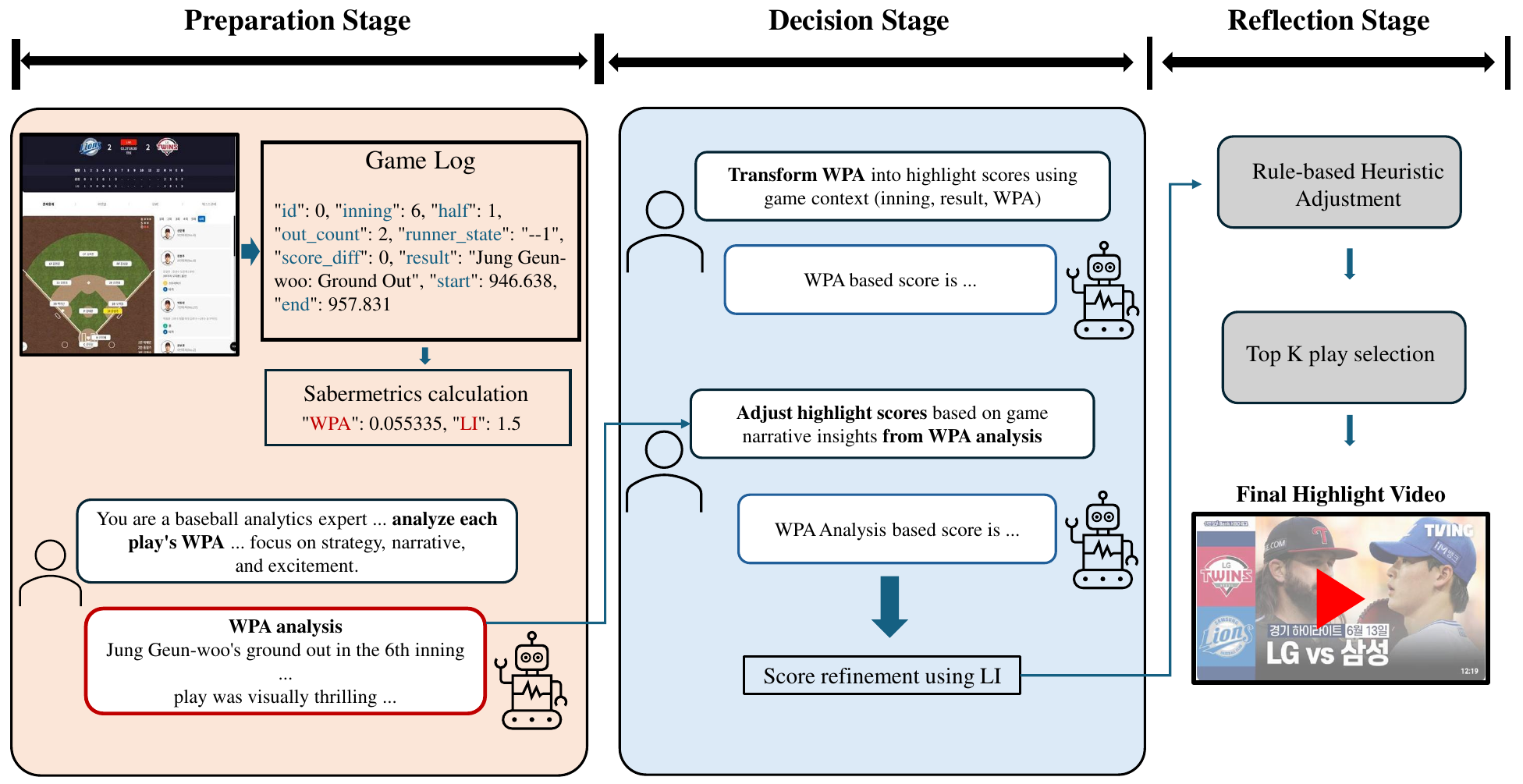}
    \caption{\textbf{The overall \ours{} framework for automated baseball highlight generation}, consisting of three stages: Preparation, Decision, and Reflection. \ours{} integrates sabermetrics with LLM-based contextual reasoning across three stages to rank, refine, and present highlights.}
    \label{fig:framework_overview}
\end{figure*}

\section{Related Works}

\subsection{Automated Sports Highlight Generation}

Automated highlight generation has been explored across baseball~\citep{naver_kbaseball, lee2020highlight}, basketball~\cite{gupta2009understanding}, soccer~\citep{decroos2017predicting}, and esports~\cite{magnifi2024, sizzle2024}, leveraging computer vision, audio analysis, and NLP to extract key moments~\cite{vasudevan2023systematic}.

\paragraph{Limitations of Traditional Methods.}
Early approaches use motion detection, scoreboard tracking~\citep{shih2017survey}, crowd noise~\citep{fu2017video}, and social signals~\cite{bettadapura2016leveraging}, but often miss contextually important plays in strategic sports like baseball~\citep{gupta2009understanding}. These methods lack interpretability and fail to model narrative flow or strategic buildup.

\paragraph{WPA-Based Approaches.}
Sabermetrics-based methods address some limitations via WPA~\citep{park2024enhancing}, which quantifies a play’s statistical impact. While effective, WPA lacks narrative nuance, often overlooking pivotal but low-WPA events (e.g., defensive setups or momentum shifts). \ours{} builds on this by incorporating contextual reasoning.

\paragraph{\ours's Contribution.}
\ours{} addresses gaps in both statistical and multimodal pipelines by integrating structured game logs with LLM-based contextual analysis. This enables transparent, language-driven highlight selection without relying on visual or audio data, making it scalable and interpretable.

\subsection{Sabermetrics and Advanced Analytics}

Sabermetrics provides structured evaluation of in-game events using metrics like \textbf{WE}, \textbf{WPA}, and \textbf{LI}~\citep{tango2007book, fangraphs_we}. These quantify situational value, outcome impact, and play criticality. While effective for analytical ranking, they fall short on narrative coherence.
\ours{} complements sabermetric precision with LLM-driven scoring to surface moments that contribute to the storyline, even when their statistical weight is low. Its modularity enables adaptation to other sports by replacing sabermetrics with domain-specific metrics (e.g., xG in soccer).

\subsection{LLMs and Agents in Sports Analytics}

\paragraph{Prior Applications of LLMs.}
LLMs have been used for structured match reports~\citep{hu2024sportsmetrics}, commentary generation~\citep{connor2023large}, and narrative templating~\citep{chiang2024badge}, but often lack integration with domain-specific numerical data. Outputs are typically rule-based and limited in adaptability.

\paragraph{Multimodal Approaches and Their Limits.}
Recent work combines video, audio, and text for event detection~\citep{della2025automated}, but these methods require rich media input and lack interpretability. In contrast, \ours{} relies on structured text data, enabling transparent, scalable highlight generation. Future work may extend \ours{} to multimodal inputs, including crowd audio and visual excitement.

\paragraph{\ours’s Integration of LLMs.}
\ours{} uses LLMs to analyze play context and sabermetric input, with prompts that encode up to five preceding plays. To reduce hallucinations, we apply constrained prompting and low-temperature decoding. Unlike prior systems, \ours{} supports explainability through traceable, structured inputs. Ablation results show that LLM-driven ranking significantly improves F1 over WPA-only baselines. By combining contextual reasoning and structured analytics, \ours{} presents a modular and extensible approach to agent-driven summarization across domains.

\section{Methodology}

\subsection{Problem Formulation}
Baseball highlight generation can be framed as an event-based summarization task, requiring the selection of key plays that balance statistical significance with narrative coherence. Traditional AI methods rely on visual/audio cues or statistical metrics like WPA, both of which fall short: the former overlooks strategic depth, and the latter ignores game momentum and storytelling. To address these gaps, we introduce \ours{}, an LLM-powered framework that integrates structured sabermetric evaluation with contextual reasoning to produce engaging, interpretable summaries.

\subsection{Framework Overview}
\ours~operates in three stages—Preparation, Decision, and Reflection (see Figure~\ref{fig:framework_overview}):

\paragraph{Preparation Stage.} Converts raw game data into structured inputs and computes sabermetric metrics (WPA, WE, LI).

\paragraph{Decision Stage.} Scores and ranks plays by combining sabermetric impact with LLM-generated contextual insights.

\paragraph{Reflection Stage.} Finalizes highlight selection by incorporating user-defined preferences and enhancing narrative coherence.

\subsection{Preparation Stage}
This stage processes raw play-by-play game logs and prepares inputs for LLM-based analysis. It includes Game Log Generation, Sabermetrics Calculation, and LLM-Based Contextual Analysis.

\paragraph{Game Log Generation.}  
We process structured game logs containing inning information, event results, and key player actions\footnote{This step involves converting raw play-by-play data from live commentary or game tracking APIs into a standardized format.}. Each key play, such as hits, strikeouts, or substitutions, is annotated with essential metadata, including timestamp, inning and half (top or bottom), result, runner state, out count, and score difference. This structured representation ensures consistency, supports sabermetric calculations, and enables reproducible, interpretable downstream analysis.

\paragraph{Sabermetrics Calculation.}  
This step computes  the sabermetric metrics WE, WPA, and LI using precomputed probability tables based on historical data. WE is derived from precomputed tables based on game-state features (inning, runner state, and score difference,  etc.). WPA measures the change in WE before and after a play, reflecting the direct impact of the event. Additionally, LI evaluates play criticality, highlighting moments with high potential for the game outcome. These metrics provide a robust quantitative foundation for ranking plays. The full definitions are provided in Appendix~\ref{app:sabermetrics_calculation}.

\paragraph{LLM-Based Contextual Analysis.}  
We use an LLM to assess play significance beyond what numerical statistics capture. The model receives structured input containing:
\begin{itemize}
    \item \textbf{Play details:} Player actions, event descriptions, and inning context.  
    \item \textbf{Recent context:} Up to 5 previous plays to capture momentum shifts.
    \item \textbf{Sabermetric:} WPA to ensure statistical grounding.
\end{itemize}

\noindent To capture the evolving narrative of a game, we adopt a sliding window approach: each play is evaluated in the context of up to five prior plays, allowing the model to reason about continuity and momentum. The LLM evaluates a play's value as a highlight based on this structured input by analyzing WPA change, game context, and visual excitement. Details on the LLM prompt are provided in Appendix~\ref{app:wpa_analysis_prompt}. By integrating quantitative metrics with qualitative insights, this step bridges statistical evaluation with human-centric storytelling, ensuring plays are assessed for both statistical significance and narrative impact.

\textbf{Example Input and Output:}
\begin{lstlisting}[language=json]
{
    "id": 35,
    "result": "Son Joo-in: Single to left field",
    "inning": "Top of the 6th",
    "WPA": -0.052,
    "previous_plays": [
        {"id": 31, "result": "Kim Min-seok: Flyout to center field"},
        {"id": 32, "result": "Park Ji-hwan: Strikeout swinging"},
        {"id": 33, "result": "Choi Jung: Walk"},
        {"id": 34, "result": "Lee Dae-ho: Single to right field"}
    ]
}
\end{lstlisting}

\begin{lstlisting}[language=json]
{
    "id": 35,
    "WPA_analysis": "This single slightly decreased the home team's chances of winning, as indicated by the negative WPA, benefiting the away team. However, given the preceding walk and another single, this sequence maintained offensive momentum, setting up potential scoring opportunities."
}
\end{lstlisting}

\subsection{Decision Stage}
The Decision Stage transforms quantitative sabermetric metrics and LLM-generated narrative insights into a unified importance score for each play.

\paragraph{WPA Transformation.}
Raw WPA values are converted into highlight scores, balancing statistical significance and game context by incorporating factors such as inning, play outcome, and overall game impact. High-leverage plays (\textit{e.g.}, a key defensive stop in the 9th inning) are weighted more heavily than a play in early innings, even with low WPA. This transformation prioritizes high-stakes moments. The LLM prompt for this process is in Appendix~\ref{app:wpa_transformation_prompt}.

\paragraph{Score Adjustment.}  
We refine scores through a two-steps:
\begin{itemize}
    \item \textit{LLM-Based Adjustments}: The LLM adjusts scores based on each play’s strategic and narrative importance using qualitative insights derived from the LLM-based contextual analysis. For plays that are strategically significant, visually exciting, or narratively crucial, their scores are increased. Pivotal plays contributing to a game’s momentum, such as key defensive stops or back-to-back scoring opportunities, receive additional weight, even if their raw WPA is relatively low.
    
    \item \textit{Leverage Index Correction}: Since WPA can undervalue non-scoring plays (\textit{e.g.}, a game-defining situation without an immediate score change), we adjust scores based on LI. Plays with high LI but low WPA receive additional weighting, with adjustments determined by the rank difference, denoted by \( \Delta R = R_{\text{WPA}} - R_{\text{LI}} \). The top-ranked plays, where LI is significantly higher than WPA, get up to 20 extra points, decreasing by 1 point per rank. For instance, the highest-ranked play receives an additional 20 points, while the second-ranked play receives 19 points, and so on. This ensures to properly value strategically significant but subtle plays.
\end{itemize}
This structured process ensures that \ours{} captures plays that matter analytically, emotionally, and contextually. Full score adjustment logic and prompts are described in Appendix~\ref{app:score_adjustment_prompt}.

\subsection{Reflection Stage}
The stage ensures highlights align with user-defined preferences and maintains narrative coherence.

\paragraph{User Preferences.}  
Our framework allows users to customize highlights based on specific preferences. For instance, users can prioritize plays from the final innings (\emph{e.g.}, 8th and 9th) or emphasize key players. These refinements ensure highlights capture both analytically significant moments and audience preferences. Additionally, scenarios such as late-inning comebacks or walk-off plays can be further emphasized to enhance strategic and emotional impact.

\paragraph{Final Highlight Selection.}  
The system automatically selects the top \( K \) plays based on adjusted scores, balancing statistical significance with narrative coherence. Users can specify the number of highlights (\( K \)) and optionally set thematic preferences (e.g., offensive or defensive plays), while the AI optimizes the selection within those constraints.

The corresponding video segments are retrieved using the timestamp data from the structured game log and enhanced with supplementary footage (\textit{e.g.}, crowd reactions, broadcast replays, and slow-motion replays of pivotal moments). The final highlight video maintains a coherent progression of events, balancing statistical significance and narrative depth for an engaging viewing experience.

\subsection{Interpretability and Modularity}
\ours{} is designed for transparency: all inputs to the LLM are structured and auditable. Scoring logic is interpretable, and the system is modular, enabling extension to other sports (e.g., soccer, basketball) or domains (e.g., esports, news). Metrics like WPA can be replaced with sport-specific features like xG or PER, and LLM components can be swapped for fine-tuned or lightweight models for real-time scenarios.

\section{Experiments and Results}

\subsection{Dataset}
We evaluate \ours{} using five KBO League games, selected based on their \textit{WE curves} to ensure a variety of game dynamics. These games include:

\begin{itemize}
    \item \textbf{Blowout games (2)}: Large score gaps, early decisive outcomes.
    \item \textbf{Close games (2)}: Outcome remained uncertain until the final innings.
    \item \textbf{Comeback game (1)}: Featured a dramatic reversal.
\end{itemize}

Due to the requirement for full-game broadcasts and verified highlight annotations, data availability was constrained. However, we curated this set to include diverse narrative arcs (e.g., momentum shifts, last-inning drama). This design offers a solid initial testbed to evaluate \ours's effectiveness across contrasting game types. Ground truth highlights were manually annotated from official broadcast highlight videos. Additional details on data sources are provided in Appendix~\ref{app:data_source}.



\subsection{Evaluation Metrics}
We evaluate using standard information retrieval and classification metrics:
\begin{itemize}
    \item \textbf{Precision}: Proportion of selected plays matching ground-truth highlights. In our context, this measures how effectively \ours{} avoids including irrelevant or unimportant plays.
    \item \textbf{Recall}: Proportion of ground-truth highlights correctly identified. This reflects \ours’s ability to capture critical narrative moments without omission.
    \item \textbf{F1-score}: Harmonic mean of Precision and Recall. This captures the overall balance between informativeness and conciseness—key goals in highlight summarization.
\end{itemize}

These metrics allow us to assess how well \ours{} identifies meaningful plays while minimizing redundancy or omission.

\subsection{Baseline Methods}
To benchmark \ours, we compare against two baselines:

\paragraph{WPA-Based Method.}  
A sabermetrics-driven approach that ranks plays solely based on WPA~\cite{park2024enhancing}, measuring their impact on a team's chances of winning. While effective in quantifying play importance, this method overlooks qualitative factors such as narrative coherence and excitement. To ensure fairness, we exclude team-specific biases commonly present in WPA-based models, generating neutral highlights for broad audience evaluation.

\paragraph{NAVER AI Highlights.}  
A proprietary video-based highlight generation system developed by NAVER~\cite{naver_kbaseball} and widely used in KBO broadcasts. While not publicly reproducible, it serves as a practical real-world reference.

\paragraph{Top-\(K\) Play Selection.}  
To maintain consistency across evaluations, we adjust the number of plays ($K$) selected by our framework and the WPA-based method such that the resulting highlight video length approximately matches that of the NAVER AI highlight baseline. For example, if NAVER’s highlight video is 10 minutes long, \(K\) is tuned accordingly to produce a highlight video of similar duration. While both \ours{} and WPA-based methods segment highlights based on structured play-by-play logs, NAVER AI Highlights often include longer and less granular sequences, affecting the total number of plays selected despite comparable video lengths.

\subsection{LLM Configuration}
We use the Mistral-Large-Instruct-2411 model~\cite{mistral2024large} for both WPA analysis and scoring, integrating sabermetrics with qualitative insights to enhance highlight generation. The model is run on four A100 GPUs via vLLM~\cite{kwon2023efficient}, enabling efficient inference while maintaining high-quality outputs. To ensure deterministic and contextually relevant results, we set the temperature to \(0\) and restrict the top-\(p\) value to \(0.1\), prioritizing the most probable completions. A 10,000-token limit prevents truncation of detailed play-by-play analyses.

\subsection{Quantitative Evaluation}
To assess the impact of different values of \( K \) (number of selected plays), we analyze F1-score variations. Results indicate that performance peaks around \( K = 60 \), balancing Precision and Recall. Accordingly, \( K \) is set per game using an experimentally determined optimal value, as detailed in Appendix~\ref{app:k_f1_analysis}.

\begin{table}[ht]
\centering
\resizebox{\columnwidth}{!}{%
\begin{tabular}{lccc}
\toprule
\textbf{Method} & \textbf{Precision} & \textbf{Recall} & \textbf{F1-score} \\
\midrule
\textcolor{gray}{\ours{}'} & \textcolor{gray}{0.814} & \textcolor{gray}{0.886} & \textcolor{gray}{0.848} \\
\ours{} (Full) & 0.748 & \textbf{0.846} & \textbf{0.793} \\
WPA-Based Method & 0.635 & 0.716 & 0.673 \\
NAVER AI Highlights & \textbf{0.818} & 0.292 & 0.429 \\
\midrule
\ours{} (– Reflection Stage) & 0.700 & 0.856 & 0.765 \\
\bottomrule
\end{tabular}%
}
\caption{\textbf{Evaluation and ablation results.} The top rows compare \ours{} with baseline methods, including a commercial system and WPA-based selection. The last row isolates the contribution of LLM-based contextual scoring by removing the Reflection Stage.}
\label{tab:combined_results}
\end{table}

\begin{table*}[ht]
\centering
\resizebox{0.7\textwidth}{!}{%
\begin{tabular}{lccccc}
\toprule
\textbf{Game Type} & \textbf{Time (m)} & \textbf{GT Time (m)} & \textbf{Total Plays} & \textbf{GT Plays} & \textbf{F1 Score} \\
\midrule
Blowout1  & 195  & 13.5  & 80  & 40  & 0.578  \\
Blowout2  & 198  & 8.3   & 82  & 54  & 0.719  \\
Close1    & 174  & 7.5   & 68  & 44  & 0.723  \\
Close2    & 233  & 9.5   & 99  & 63  & 0.842  \\
Comeback  & 185  & 10.3  & 84  & 47  & 0.680  \\
\bottomrule
\end{tabular}%
}
\caption{\textbf{Performance of~\ours~across different game types}. 
Time refers to the total broadcast video length of the game in minutes, 
GT Time represents the length of the official highlight video in minutes, 
Total Plays is the number of plays in the game, and
GT Plays indicates the number of ground truth plays from the official highlight video.}
\label{tab:game_performance}
\end{table*}

Table~\ref{tab:combined_results} summarizes the results. NAVER AI Highlights achieves the highest Precision (0.818) due to its selective approach, focusing on a limited number of plays. However, its Recall (0.292) is the lowest, as it omits a wide range of critical moments, leading to an overall F1-score of 0.429. The method prioritizes scoring plays, resulting in longer but less granular highlight segments.

In contrast, the WPA-based method achieves a higher Recall (0.716) by capturing a broader range of plays, but its lower Precision (0.635) suggests it includes many marginally relevant events. This reflects its limitation in filtering for narrative coherence.

Our framework \ours{} achieves the best balance between Precision (0.814) and Recall (0.886), resulting in the highest F1-score (0.848) on overlapping games (\textcolor{gray}{\ours{}'}) and 0.793 across all games (\ours{}). This demonstrates its ability to effectively capture key moments while maintaining contextual integrity.

\paragraph{Impact of Game Type on Performance.}  
We further analyze performance across different game types (Table~\ref{tab:game_performance}). Close games tend to yield higher F1-scores, likely due to frequent pivotal moments. While comeback games were expected to perform well due to their dramatic nature, results suggest that WPA-based rankings may undervalue early plays that contribute to later momentum shifts. Blowout games exhibit mixed results, as the highlight-worthiness of individual plays is influenced by isolated high-impact moments rather than overall game tension.

Additionally, while longer games naturally contain more plays, our results do not indicate a direct correlation between game length and highlight selection performance. Instead, the distribution of highlights and the diversity of events appear to be more influential in determining the F1 score. More studies with a larger dataset will be necessary to confirm these observations.

\subsection{Qualitative Evaluation}
We conducted a user study with three expert curators from TVING\footnote{\url{https://www.tving.com/}}, all experienced in professional sports media production. Participants watched paired highlight videos (\ours{} vs. NAVER-AI) and evaluated \textbf{key moment coverage, narrative coherence, scene diversity, and informativeness}, selecting a preferred version or indicating no preference (tie = 50\%).

While the participant pool is small, the use of domain experts provides high-quality qualitative feedback in the absence of standardized storytelling metrics. Future work will expand the participant set and include inter-annotator agreement scores.

\begin{table}[ht]
\centering
\resizebox{\columnwidth}{!}{%
\begin{tabular}{l c}
\toprule
\textbf{Evaluation Criteria} & \textbf{Preference for \ours{} (\%)} \\
\midrule
Key Moment Coverage & 50.0 \\
Narrative Coherence & 66.6 \\
Scene Diversity & 66.6 \\
Informativeness & 66.6 \\
Overall Satisfaction & 66.6 \\
\bottomrule
\end{tabular}%
}
\caption{\textbf{User study results.} Preference (\%) indicates the proportion of participants favoring \ours, with ties recorded as 50\%. Expert curators preferred \ours{} across most dimensions.}
\label{tab:qualitative_results}
\end{table}

\paragraph{Results and Analysis.}  
\ours~was preferred over NAVER-AI in 4 of 5 qualitative evaluation aspects, indicating its effectiveness in generating engaging and comprehensive highlights. In particular,

\begin{enumerate}
    \item \textbf{Key Moment Coverage}: Preferences were evenly split (50\%), suggesting both systems captured crucial plays equally well.
    \item \textbf{Narrative Coherence}: Despite NAVER-AI’s potentially more structured flow, 66.6\% of participants preferred \ours, indicating a more engaging and dynamic storytelling experience.
    \item \textbf{Scene Diversity and Informativeness}: \ours~was favored by 66.6\% of participants in both aspects, highlighting its ability to showcase a broader range of plays and provide richer game context.
    \item \textbf{Overall Satisfaction}: \ours~was the preferred choice for 66.6\% of participants, reinforcing its ability to create an immersive and memorable highlight experience.
\end{enumerate}
While NAVER-AI maintains a more rigid scene structure, these results suggest \ours’s storytelling advantages are recognized by expert users, despite the small study size. The highlight videos used in the user study are provided in Appendix~\ref{app:user_study_videos}.

\subsection{Ablation Study}
To isolate the contributions of individual components, we compare the following configurations:
\begin{enumerate}
    \item WPA-only baseline  
    \item \ours{} without the Reflection Stage (i.e., Preparation + Decision only)  
    \item Full \ours{} pipeline
\end{enumerate}
The results are presented in Table~\ref{tab:combined_results}. The 9.2-point F1 improvement from WPA to \ours(– Reflection) validates the effectiveness of the Preparation and Decision Stages in identifying key plays through sabermetric-driven selection and LLM-enhanced analysis. For consistency, we set \( k = 60 \) as the selection threshold. An additional 2.8-point gain is observed with the full pipeline, confirming the value of user preference integration and narrative refinement. While \ours(– Reflection) excludes final-inning prioritization and thematic tuning, it still significantly outperforms WPA-based methods, highlighting the strength of \ours’s structured, interpretable decision-making process.

\section{Conclusion}


We introduced \ours{}, an LLM-driven agent for context-aware baseball highlight summarization. By integrating structured sabermetric metrics with narrative-aware LLM reasoning, \ours{} balances statistical rigor with storytelling coherence—without relying on multimodal inputs. Despite being evaluated on a small but diverse set of KBO games, \ours{} outperforms both commercial and statistical baselines. Ablation studies and expert evaluations highlight the effectiveness of its modular agent pipeline, with measurable gains from contextual scoring and user-guided refinement. Future work will explore extensions to other sports and domains—for example, adapting the framework using expected goals (xG) in soccer or player efficiency rating (PER) in basketball—as well as multimodal integration and lightweight LLMs for real-time applications.





\section{Limitations}

While \ours{} demonstrates strong performance in automated baseball highlight generation, several limitations warrant further investigation and refinement.

\paragraph{Dataset scale and coverage.}
Our evaluation is based on five full-length KBO games, selected to reflect diverse game dynamics including blowouts, comebacks, and close contests. While this diversity helps validate the framework under varied scenarios, the dataset remains limited in scale due to restricted access to broadcast footage and the difficulty of acquiring verified ground-truth highlights. Broader validation across additional leagues (e.g., MLB, NPB), game formats, and sports will be necessary to assess generalizability.

\paragraph{User study limitations.}
Our qualitative evaluation was conducted with three domain experts. While their professional experience supports the validity of feedback, the small participant size limits generalizability. Future work will involve a broader and more diverse pool of evaluators to improve reliability and coverage.

\paragraph{Reliance on WPA-based significance evaluation.}  
Although WPA and related metrics offer a solid foundation for statistical scoring, they may undervalue early plays that set up later turning points. While the LLM mitigates this by reasoning over recent play sequences, long-term momentum modeling remains a challenge. Integrating sequence-aware models or play-level memory mechanisms could help.

\paragraph{Absence of multimodal signals.}  
\ours{} relies solely on structured text and sabermetric input, omitting visual and auditory cues (e.g., crowd reactions, broadcast replays). These can capture excitement and atmosphere, which are important in highlight curation. We plan to incorporate multimodal features in future work while maintaining \ours’s interpretability.

\paragraph{LLM hallucination risk and lack of domain adaptation.}  
We use a general-purpose LLM without fine-tuning. While prompt constraints and low-temperature decoding reduce hallucinations, some outputs may still misrepresent context or overemphasize certain play types. Future iterations will explore domain-specific fine-tuning, explanation tracing, and bias mitigation.

\paragraph{Manual heuristics in user preference integration.}  
User customization is currently based on configurable heuristic rules in the Reflection Stage. While this supports interpretability, it limits adaptability. We plan to develop learning-based re-ranking strategies to support more robust, data-driven personalization.

\paragraph{Real-time applicability and efficiency.}  
\ours{} is designed for post-game summarization. Although runtime is practical due to structured input and parallelized inference, real-time deployment remains a challenge. Future work will investigate lightweight LLMs, prompt distillation, and streaming pipelines for live scenarios.

\paragraph{User study limitations.}
Our qualitative evaluation was conducted with three domain experts. While their professional experience supports the validity of feedback, the small participant size limits generalizability. Future work will involve a broader pool of evaluators to improve reliability and coverage.

Despite these limitations, \ours{} demonstrates the feasibility of interpretable, modular highlight summarization using structured data and contextual reasoning. Ongoing research will extend the framework’s robustness, efficiency, and applicability across domains.

\section*{Acknowledgments}
We thank our colleagues at the AI R\&D Division for their valuable feedback and discussions during the development of this work. We also thank the team at TVING for supporting the human evaluation and contributing expert insights during testing. 

Joonseok Lee was supported by NRF grants (RS-2021-NR05515, RS-2024-00336576) and IITP grants (RS-2024-00353131, RS-2022-II220264), funded by the Korean government.

\bibliography{references}

\begin{thebibliography}{23}
\providecommand{\natexlab}[1]{#1}

\bibitem[{Bettadapura et~al.(2016)Bettadapura, Pantofaru, and Essa}]{bettadapura2016leveraging}
Vinay Bettadapura, Caroline Pantofaru, and Irfan Essa. 2016.
\newblock Leveraging contextual cues for generating basketball highlights.
\newblock In \emph{Proceedings of the 24th ACM international conference on Multimedia}, pages 908--917.

\bibitem[{Chiang et~al.(2024)Chiang, Chao, Wang, Wang, and Peng}]{chiang2024badge}
Shang-Hsuan Chiang, Lin-Wei Chao, Kuang-Da Wang, Chih-Chuan Wang, and Wen-Chih Peng. 2024.
\newblock Badge: Badminton report generation and evaluation with llm.
\newblock \emph{arXiv preprint arXiv:2406.18116}.

\bibitem[{Connor and O'Neill(2023)}]{connor2023large}
Mark Connor and Michael O'Neill. 2023.
\newblock Large language models in sport science \& medicine: Opportunities, risks and considerations.
\newblock \emph{arXiv preprint arXiv:2305.03851}.

\bibitem[{Decroos et~al.(2017)Decroos, Dzyuba, Van~Haaren, and Davis}]{decroos2017predicting}
Tom Decroos, Vladimir Dzyuba, Jan Van~Haaren, and Jesse Davis. 2017.
\newblock Predicting soccer highlights from spatio-temporal match event streams.
\newblock In \emph{Proceedings of the AAAI Conference on Artificial Intelligence}, volume~31.

\bibitem[{Della~Santa and Lalli(2025)}]{della2025automated}
Francesco Della~Santa and Morgana Lalli. 2025.
\newblock Automated detection of sport highlights from audio and video sources.
\newblock \emph{arXiv preprint arXiv:2501.16100}.

\bibitem[{{FanGraphs}(2025)}]{fangraphs_we}
{FanGraphs}. 2025.
\newblock \href {https://library.fangraphs.com/misc/we/win-expectancy/} {https://library.fangraphs.com}.

\bibitem[{Fu et~al.(2017)Fu, Lee, Bansal, and Berg}]{fu2017video}
Cheng-Yang Fu, Joon Lee, Mohit Bansal, and Alexander~C Berg. 2017.
\newblock Video highlight prediction using audience chat reactions.
\newblock \emph{arXiv preprint arXiv:1707.08559}.

\bibitem[{Gupta et~al.(2009)Gupta, Srinivasan, Shi, and Davis}]{gupta2009understanding}
Abhinav Gupta, Praveen Srinivasan, Jianbo Shi, and Larry~S Davis. 2009.
\newblock Understanding videos, constructing plots learning a visually grounded storyline model from annotated videos.
\newblock In \emph{2009 IEEE Conference on Computer Vision and Pattern Recognition}, pages 2012--2019. IEEE.

\bibitem[{Hu et~al.(2024)Hu, Song, Cho, Wang, Foroosh, Yu, and Liu}]{hu2024sportsmetrics}
Yebowen Hu, Kaiqiang Song, Sangwoo Cho, Xiaoyang Wang, Hassan Foroosh, Dong Yu, and Fei Liu. 2024.
\newblock Sportsmetrics: Blending text and numerical data to understand information fusion in llms.
\newblock \emph{arXiv preprint arXiv:2402.10979}.

\bibitem[{Jiang et~al.(2020)Jiang, Qu, Wang, Wang, and Zheng}]{jiang2020towards}
Ruochen Jiang, Changbo Qu, Jiannan Wang, Chi Wang, and Yudian Zheng. 2020.
\newblock Towards extracting highlights from recorded live videos: An implicit crowdsourcing approach.
\newblock In \emph{2020 IEEE 36th International Conference on Data Engineering (ICDE)}, pages 1810--1813. IEEE.

\bibitem[{Joshi et~al.(2017)Joshi, Merler, Nguyen, Hammer, Kent, Smith, and Feris}]{joshi2017ibm}
Dhiraj Joshi, Michele Merler, Quoc-Bao Nguyen, Stephen Hammer, John Kent, John~R Smith, and Rogerio~S Feris. 2017.
\newblock Ibm high-five: Highlights from intelligent video engine.
\newblock In \emph{Proceedings of the 25th ACM international conference on Multimedia}, pages 1249--1250.

\bibitem[{Kwon et~al.(2023)Kwon, Li, Zhuang, Sheng, Zheng, Yu, Gonzalez, Zhang, and Stoica}]{kwon2023efficient}
Woosuk Kwon, Zhuohan Li, Siyuan Zhuang, Ying Sheng, Lianmin Zheng, Cody~Hao Yu, Joseph Gonzalez, Hao Zhang, and Ion Stoica. 2023.
\newblock Efficient memory management for large language model serving with pagedattention.
\newblock In \emph{Proceedings of the 29th Symposium on Operating Systems Principles}, pages 611--626.

\bibitem[{Lee et~al.(2020)Lee, Jung, Yang, and Lee}]{lee2020highlight}
Younghyun Lee, Hyunjo Jung, Cheoljong Yang, and Joonsoo Lee. 2020.
\newblock Highlight-video generation system for baseball games.
\newblock In \emph{2020 IEEE International Conference on Consumer Electronics-Asia (ICCE-Asia)}, pages 1--4. IEEE.

\bibitem[{Magnifi(2025)}]{magnifi2024}
Magnifi. 2025.
\newblock \href {https://magnifi.ai} {https://magnifi.ai}.

\bibitem[{Merler et~al.(2018)Merler, Mac, Joshi, Nguyen, Hammer, Kent, Xiong, Do, Smith, and Feris}]{merler2018automatic}
Michele Merler, Khoi-Nguyen~C Mac, Dhiraj Joshi, Quoc-Bao Nguyen, Stephen Hammer, John Kent, Jinjun Xiong, Minh~N Do, John~R Smith, and Rogerio~Schmidt Feris. 2018.
\newblock Automatic curation of sports highlights using multimodal excitement features.
\newblock \emph{IEEE Transactions on Multimedia}, 21(5):1147--1160.

\bibitem[{MistralAI(2024)}]{mistral2024large}
MistralAI. 2024.
\newblock \href {https://huggingface.co/mistralai/Mistral-Large-Instruct-2411} {Mistral-large-instruct-2411}.

\bibitem[{NAVER(2025)}]{naver_kbaseball}
NAVER. 2025.
\newblock \href {https://tv.naver.com/kbaseball} {https://tv.naver.com/kbaseball}.

\bibitem[{Park et~al.(2024)Park, Lim, Lee, and Suh}]{park2024enhancing}
Kieun Park, Hajin Lim, Joonhwan Lee, and Bongwon Suh. 2024.
\newblock Enhancing auto-generated baseball highlights via win probability and bias injection method.
\newblock In \emph{Proceedings of the CHI Conference on Human Factors in Computing Systems}, pages 1--18.

\bibitem[{{Retrosheet}(2025)}]{retrosheet}
{Retrosheet}. 2025.
\newblock \href {https://www.retrosheet.org/} {https://www.retrosheet.org}.

\bibitem[{Shih(2017)}]{shih2017survey}
Huang-Chia Shih. 2017.
\newblock A survey of content-aware video analysis for sports.
\newblock \emph{IEEE Transactions on circuits and systems for video technology}, 28(5):1212--1231.

\bibitem[{Sizzle(2025)}]{sizzle2024}
Sizzle. 2025.
\newblock \href {https://www.sizzlehighlights.com} {https://www.sizzlehighlights.com}.

\bibitem[{Tango et~al.(2007)Tango, Lichtman, and Dolphin}]{tango2007book}
Tom~M Tango, Mitchel~G Lichtman, and Andrew~E Dolphin. 2007.
\newblock \emph{The book: Playing the percentages in baseball}.
\newblock Potomac Books, Inc.

\bibitem[{Vasudevan and Gounder(2023)}]{vasudevan2023systematic}
Vani Vasudevan and Mohan~S Gounder. 2023.
\newblock A systematic review on machine learning-based sports video summarization techniques.
\newblock \emph{Smart Computer Vision}, pages 1--34.

\end{thebibliography}

\clearpage
\twocolumn 
\appendix
\section*{Appendix}
\addcontentsline{toc}{section}{Appendix}

\section{Sabermetrics Calculation}
\label{app:sabermetrics_calculation}

\subsection{Win Expectancy (WE)}
WE represents the probability of a team winning the game given the current game state. It is computed using historical game data:

\begin{equation}
    WE(s) = \frac{W_s}{N_s}
\end{equation}

where:
\begin{itemize}
    \item $WE(s)$: Win Expectancy at state $s$
    \item $W_s$: Number of games won from state $s$
    \item $N_s$: Total number of games observed in state $s$
\end{itemize}

Rather than computing WE dynamically, we utilize a precomputed \textbf{Win Expectancy Table} derived from historical game outcomes.

\subsection{Win Probability Added (WPA)}
WPA quantifies the impact of a specific play on a team's win probability:

\begin{equation}
    WPA = WE_{\text{after}} - WE_{\text{before}}
\end{equation}

where:
\begin{itemize}
    \item $WE_{\text{before}}$: Win Expectancy before the play
    \item $WE_{\text{after}}$: Win Expectancy after the play
\end{itemize}

A positive WPA indicates an improvement in the team's win probability, while a negative WPA signifies a decrease.

\subsection{Leverage Index (LI)}
LI measures the potential significance of a game situation based on the expected shift in WE:

\begin{equation}
    LI = \frac{|WE_{\text{after}} - WE_{\text{before}}|}{\text{Avg}(|WE_{\text{after}} - WE_{\text{before}}|)}
\end{equation}

where:
\begin{itemize}
    \item $WE_{\text{before}}$: Win Expectancy before the play
    \item $WE_{\text{after}}$: Win Expectancy after the play
    \item $\text{Avg}(|WE_{\text{after}} - WE_{\text{before}}|)$: Average absolute WE change across all historical plays
\end{itemize}

A higher LI value indicates a high-stakes moment where a play has greater potential to influence the game's outcome.

\clearpage
\onecolumn

\section{LLM Prompts Used in \ours{}}
\subsection{Prompt for WPA Analysis}
\label{app:wpa_analysis_prompt}
\framebox{%
\parbox{\dimexpr\textwidth-2\fboxsep-2\fboxrule\relax}{%
\setlength{\parskip}{1em} 
\raggedright 
You are a baseball analytics expert. Your task is to analyze and evaluate each play of a baseball game based on the given input data, which includes commentary, the result of the play, the inning, and the calculated Win Probability Added (WPA). Your goal is to provide a detailed analysis of each play's significance in terms of the overall game narrative, its potential for visual excitement, and how it fits into the game's context. Follow these instructions closely:

\textbf{1. Task Overview:}
You are provided with data for key plays in the game. Each play has an ID, a result (e.g., a hit or an out), the inning it occurred, and a WPA value. Note that the WPA value is calculated based on the home team's win probability. Therefore:
\begin{itemize}
    \item A positive WPA value indicates a play that improved the home team's chances of winning.
    \item A negative WPA value suggests that the play benefited the away team, increasing their chances of winning.
\end{itemize}

\textbf{2. Instructions for Analyzing Each Play:}
\begin{itemize}
    \item \textbf{WPA Importance:} Evaluate the significance of the WPA value. Assess how significant the change in WPA is relative to the overall game.
    \item \textbf{Game Context:} Consider when the play occurred (inning) and which team was on the offensive. Was it a crucial moment? Did it help or hurt the team's chances of winning?
    \item \textbf{Visual Excitement:} Assess whether the play had potential for generating excitement from the fans (e.g., a clutch hit or a defensive gem).
    \item \textbf{Narrative Impact:} Evaluate how the play contributes to the broader story of the game.
\end{itemize}

\textbf{3. Additional Considerations:}
\begin{itemize}
    \item \textbf{Clutch Moments:} If a play occurs late in the game or in a high-leverage situation, emphasize its importance in shaping the final outcome.
    \item \textbf{Narrative Shifts:} If a play significantly shifts momentum (either by boosting or hurting a team's chances), highlight this in your analysis.
    \item \textbf{Low-Impact Plays:} If the WPA change is minimal, you can note that while the play may not be dramatic, it contributed to the ebb and flow of the game.
    \item \textbf{Varied Language:} Use diverse language and descriptors to make each analysis unique and engaging.
    \item \textbf{Comparative Analysis:} When relevant, reference previous significant plays to highlight shifts in momentum or game narrative.
    \item \textbf{Team-Specific Context:} Keep in mind that a negative WPA during an away team's offensive play can have positive implications for the away team.
\end{itemize}
}}
\twocolumn 

\subsection{Prompt for WPA Transformation}
\label{app:wpa_transformation_prompt}
\framebox{%
\parbox{\dimexpr\textwidth-2\fboxsep-2\fboxrule\relax}{%
\setlength{\parskip}{1em} 
\raggedright 
You are HighlightAI, an advanced assistant tasked with scoring the importance of each play in a baseball game based on its statistical impact using Win Probability Added (WPA), inning, and result. Your goal is to evaluate each play and assign scores from 1 to 60, focusing on the overall impact of the play on the game.

\textbf{Input Data:}
For each play, you will receive the following fields:
\begin{itemize}
    \item \texttt{"id"}: The unique identifier for the play.
    \item \texttt{"result"}: The result of the play (e.g., hit, out, home run).
    \item \texttt{"inning info"}: The inning and half-inning when the play occurred.
    \item \texttt{"WPA"}: The Win Probability Added value, which measures how much the play affected the team's chances of winning.
\end{itemize}

\textbf{Scoring Process:}
Assign a score between 1 and 60 to each play based on its WPA (Win Probability Added) value, inning, and result, which together measure the statistical and situational impact of the play on the game’s outcome.

\textbf{Updated Scoring Guidelines:}
\begin{itemize}
    \item \textbf{High-Impact Moments (40-60):}
    Assign these scores to plays with a high absolute WPA value (\( \geq 0.15 \)) and that occur in crucial moments (e.g., late innings or key situations like a home run or game-saving play).

    \item \textbf{Moderate-Impact Moments (20-39):}
    Assign these scores to plays with a moderate absolute WPA value (\( 0.05 \leq \text{WPA} < 0.15 \)), or plays that occur in mid-game but still influence momentum.

    \item \textbf{Low-Impact Moments (1-19):}
    Assign these scores to plays with a low absolute WPA value (\( < 0.05 \)), typically routine plays or early-game moments with minimal influence on the overall outcome.
\end{itemize}

\textbf{Response Format:}
Return the results as a JSON array where each play has:
\begin{itemize}
    \item \texttt{"id"}: The unique identifier for the play.
    \item \texttt{"score"}: An integer between 1 and 60, reflecting the play's importance based on WPA, inning, and result.
    \item \texttt{"rationale"}: A brief explanation for the score, mentioning key factors (e.g., WPA, inning, result).
\end{itemize}

\textbf{Scoring Focus:}
Assign scores based only on the WPA value, inning, and result. No other factors, such as commentary or narrative, should be considered at this stage.
}}

\onecolumn

\subsection{Prompt for Score Adjustment using WPA Analysis}
\label{app:score_adjustment_prompt}
\framebox{%
\parbox{\dimexpr\textwidth-2\fboxsep-2\fboxrule\relax}{%
\setlength{\parskip}{1em} 
\raggedright 
You are HighlightAI, an advanced assistant tasked with scoring the importance of each play in a baseball game based on both its statistical impact and strategic context. Your goal is to refine the score for each play by using both the score (calculated from the WPA) and the WPA analysis (which provides additional strategic insight).

\textbf{Input Data:}
For each play, you will receive the following fields:
\begin{itemize}
    \item \texttt{"id"}: The unique identifier for the play.
    \item \texttt{"result"}: The result of the play (e.g., hit, out, home run).
    \item \texttt{"inning info"}: The inning and half-inning when the play occurred.
    \item \texttt{"score"}: The initial score based on the Win Probability Added (WPA) value, which reflects the statistical impact of the play.
    \item \texttt{"WPA analysis"}: A brief analysis that provides strategic insight into the play, explaining how the play influenced the game or momentum.
\end{itemize}

\textbf{Scoring Process:}
Refine the score for each play by adjusting the score based on the insights provided in the WPA analysis. Increase the score by a minimum of +1 to a maximum of +20 to reflect both the statistical impact of the play and the strategic significance highlighted in the WPA analysis.

\textbf{Scoring Guidelines:}
\begin{itemize}
    \item \textbf{Significant Strategic Impact (+10 to +20):}
    If the WPA analysis highlights key strategic decisions, momentum shifts, or critical game-defining moments, increase the score significantly.
    
    \item \textbf{Moderate Strategic Impact (+5 to +10):}
    If the WPA analysis indicates a notable but less game-changing influence on the play’s significance, increase the score moderately.

    \item \textbf{Minimal Strategic Impact (+1 to +5):}
    If the WPA analysis suggests only minor strategic importance or has little additional significance beyond the score, increase the score slightly.
\end{itemize}

\textbf{Response Format:}
Return the results as a JSON array where each play has:
\begin{itemize}
    \item \texttt{"id"}: The unique identifier for the play.
    \item \texttt{"score"}: An integer number, reflecting the refined importance of the play.
    \item \texttt{"rationale"}: A brief explanation for the score, mentioning both the score and any adjustments made based on the WPA analysis.
\end{itemize}

\textbf{Scoring Focus:}
Use both the score and the strategic insights from WPA analysis to adjust and finalize the score. Provide a rationale that explains the adjustment. Ensure that:
\begin{itemize}
    \item No negative score adjustments are made.
    \item All scores increase by at least +1, regardless of the strategic impact described in WPA analysis.
\end{itemize}
}}

\clearpage
\twocolumn

\section{Dataset Sources}
\label{app:data_source}

To evaluate our framework, we utilized game footage and highlights from various sources. The dataset includes full game videos, ground-truth (GT) highlights, and NAVER AI-generated highlights where available.

\begin{itemize}
    \item \textbf{20230616 Doosan Bears vs. LG Twins}
        \begin{itemize}
            \item Full Game Video: \url{https://m.sports.naver.com/video/1080292}
            \item GT Highlights: \url{https://m.sports.naver.com/video/1080301}
            \item NAVER AI Highlights: \url{https://m.sports.naver.com/video/1080286}
        \end{itemize}
        
    \item \textbf{20230919 SSG Landers vs. Hanwha Eagles}
        \begin{itemize}
            \item Full Game Video: \url{https://m.sports.naver.com/video/1110228}
            \item GT Highlights: \url{https://m.sports.naver.com/video/1110203}
            \item NAVER AI Highlights: \url{https://m.sports.naver.com/video/1110184}
        \end{itemize}
        
    \item \textbf{20240925 Lotte Giants vs. Kia Tigers}
        \begin{itemize}
            \item Full Game Video: TVING Provided
            \item GT Highlights: \url{https://www.youtube.com/@tvingsports}
        \end{itemize}
        
    \item \textbf{20160409 Hanwha Eagles vs. NC Dinos}
        \begin{itemize}
            \item Full Game Video: \url{https://m.sports.naver.com/video/185165}
            \item GT Highlights: \url{https://m.sports.naver.com/video/185146}
        \end{itemize}
        
    \item \textbf{20160825 SK Wyverns vs. KT Wiz}
        \begin{itemize}
            \item Full Game Video: \url{https://m.sports.naver.com/video/230870}
            \item GT Highlights: \url{https://m.sports.naver.com/video/230861}
        \end{itemize}
\end{itemize}




\section{Optimizing Top \( K \) Selection for Highlight Generation}
\label{app:k_f1_analysis}

\begin{figure}[h]
\centering
\includegraphics[width=\columnwidth]{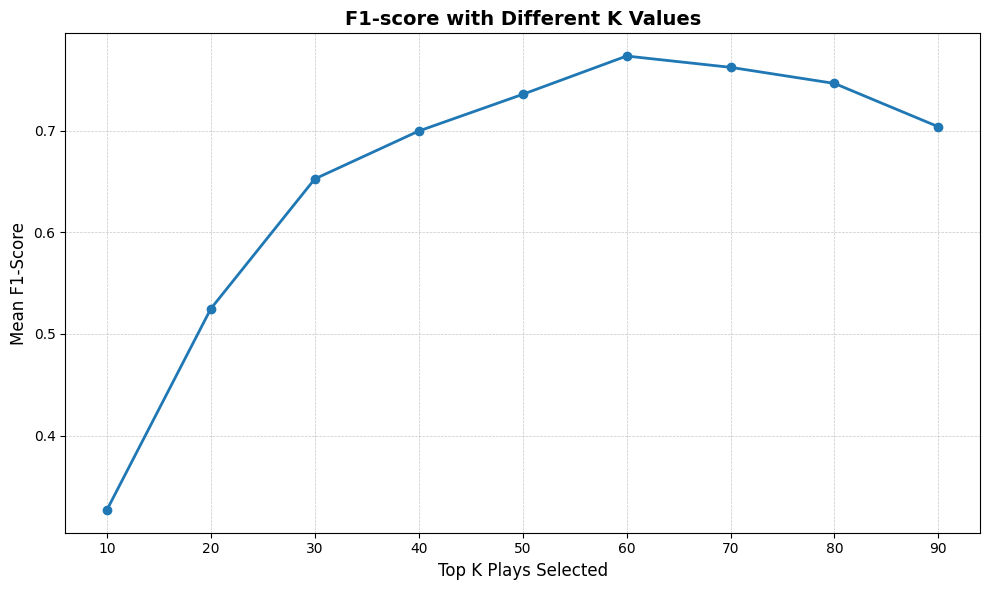}
\caption{Mean F1-score across different top \( K \) values. The score peaks around \( K = 60 \), balancing recall and precision.}
\label{fig:top_k_f1_appendix}
\end{figure}

To evaluate the impact of the top \( K \) selection on performance, we varied \( K \) from 10 to 90 in increments of 10 and computed the mean F1-score across all games. As shown in Figure~\ref{fig:top_k_f1_appendix}, the F1-score peaks around \( K = 60 \), indicating that selecting too many plays reduces precision, while selecting too few lowers recall.


\section{User Study: Highlight Videos}
\label{app:user_study_videos}

The highlight videos evaluated in the user study are provided below:

\begin{itemize}
    \item \textbf{\ours{} Highlights}: \url{https://vimeo.com/1049947443/691bd5c761}
    \item \textbf{NAVER AI Highlights}: \url{https://vimeo.com/1049947397/5fb34ab552}
\end{itemize}

These videos showcase differences in highlight generation approaches, highlighting variations in statistical analysis, narrative coherence, and audience engagement.

\end{document}